\documentclass[letterpaper, 10 pt, conference]{ieeeconf}  

\IEEEoverridecommandlockouts                              

\usepackage{times} 

\usepackage{graphicx}
\usepackage{comment}
\usepackage{hyperref}
\usepackage{hyperref}
\usepackage{multirow}
\usepackage{xcolor}
\usepackage{amsmath}
\usepackage{amsfonts}
\usepackage{balance}
\usepackage[space]{cite}

\title{\LARGE \bf
VERAGMIL: Virtual Environment for Scooping Granular Foods with Imitation Learning Models
}
\author{Amanuel Ergogo$^{1,3}$, Diego Dall’Alba$^{1,2}$ and Przemyslaw Korzeniowski$^{1}$
\thanks{The publication was created within the project of the Minister of Science and Higher Education "Support for the activity of Centers of Excellence established in Poland under Horizon 2020" on the basis of the contract number MEiN/2023/DIR/3796. This project has received funding from the EU’s Horizon 2020 research and innovation programme under grant agreement No 857533. This publication is supported by Sano project carried out within the International Research Agendas programme of the Foundation for Polish Science, co-financed by the EU Regional Development Fund.}
\thanks{$^{1}$ SANO Centre for Computational Personalized Medicine, Krakow, Poland.
        {\tt\small amanuel.ergogo@gmail.com}}%
\thanks{$^{2}$ Department of Engineering for Innovation Medicine, University of Verona, Verona, Italy.
        {\tt\small }}%
\thanks{$^{3}$ Department of Computer Science and Engineering, University of South Florida, Tampa, United States.
        {\tt\small aergogo@usf.edu}}%
}

\begin{document}

\maketitle
\begingroup
\renewcommand{\thefootnote}{}
\footnotetext{%
© 2025 IEEE. Personal use of this material is permitted. Permission from IEEE must be obtained for all other uses, in any current or future media, including reprinting/republishing this material for advertising or promotional purposes, creating new collective works, for resale or redistribution to servers or lists, or reuse of any copyrighted component of this work in other works.
DOI: \url{https://doi.org/10.1109/IROS60139.2025.11247362}
}
\endgroup
\thispagestyle{empty}
\pagestyle{empty}

\begin{abstract}
    Robot-Assisted Feeding (RAF) systems are  essential for assisting individuals with disabilities or motor impairments in eating tasks. Manipulating granular food items, such as rice and beans, poses significant challenges due to their dynamic physical properties. Learning from human demonstrations offers a promising solution, but acquiring high-quality demonstrations is complex. To address this, we present VERAGMIL, a framework that combines a high-fidelity simulator with an intuitive Virtual Reality (VR) interface for recording demonstrations and supporting different imitation learning methods. VERAGMIL provides a realistic environment for training RAF systems to handle granular materials, including robots, sensors, and various food items with distinct physical characteristics. We evaluate VERAGMIL by training three imitation learning models—BC, BC-RNN, and BCQ—on granular scooping and transporting tasks using both VR interface and 3D space mouse demonstrations, comparing them with a human-expert baseline. The models are assessed on success rate, spillage, generalization to unseen food items, and task completion time. Results show that VR-based demonstrations significantly outperform 3D space mouse data, with BCQ achieving the best overall performance, particularly in reducing spillage and approaching human performance. These findings underscore the effectiveness of our framework for training RAF systems in granular material handling. The code for our framework is publicly available at: \href{https://github.com/AmanuelErgogo/VERAGMIL.git}{\texttt{https://github.com/AmanuelErgogo/VERAGMIL.git}}.
\end{abstract}

\section{INTRODUCTION}
 
Due to recent advancements in computation, sensing, and hardware, robots are increasingly being adopted in human environments \cite{nanavati2023, lancioni2021}. Assistive robots, in particular, help individuals with daily tasks \cite{sarath2022, yamazaki2021}. 
Robot-Assisted Feeding (RAF) systems are an essential component of assistive robotics, providing critical support for individuals with disabilities, the elderly, and those with motor impairments to eat independently \cite{bhattacharjee2020,erickson2019}. Since eating is one of the most challenging tasks for these users, RAF boosts their sense of independence and significantly reduces the time caregivers spend on feeding, which is often one of the most time-consuming caregiving tasks \cite{park2019,gordon2020}. RAF systems assist in scooping, transporting, and transferring food \cite{sundaresan2022, grannen2022}. While rigid and semi-solid foods are well-handled, granular materials like rice and pasta remain challenging due to their unpredictable behavior \cite{gordon2019,sundaresan2022}. Precision and reliability are critical in these tasks, but existing RAF systems that excel with rigid foods struggle with the complexities of handling loose, particulate materials.

\begin{figure}[t]
     \centering
     \includegraphics[width=0.45\textwidth]{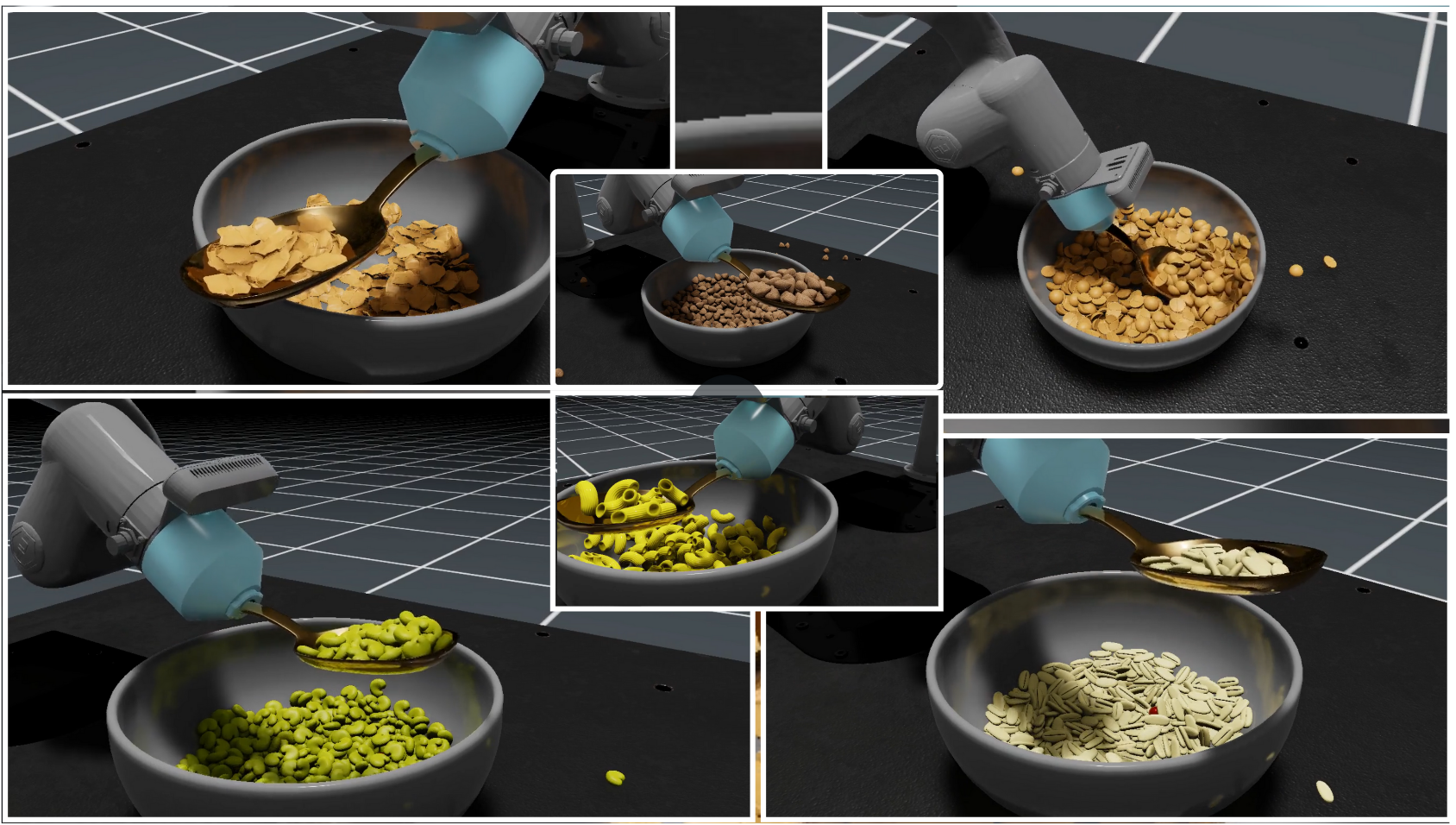}
     \caption{VERAGMIL environment showing robotic arm (\textit{Ufactory xArm7}) scooping and transporting various granular food items with different shapes, sizes, and physical properties, demonstrating adaptive manipulation of our robot-assisted feeding system.}
     \label{fig:gran-mat}
 \end{figure}

The difficulty in manipulating granular materials arises from their ability to behave both as solids and as fluids, depending on external forces \cite{nanavati2023,parikh2023}. 
In contrast to rigid food items, which respond predictably to manipulation, granular materials exhibit nonlinear behavior, making them more challenging to handle for robotic systems \cite{park2017}. These problems are particularly pronounced in RAF, where even minor inaccuracies can lead to significant spillage, reducing the reliability and usability of the system 
\cite{gordon2019,parikh2023}.

Previous research in RAF has focused mainly on handling rigid or semi-solid foods, leaving the challenge of granular material manipulation relatively underexplored \cite{bhattacharjee2020,park2019}. 
Techniques that work well with rigid food items do not translate effectively to granular foods, where dynamic control and adaptability are necessary \cite{sundaresan2022,erickson2020}. Due to the complex and highly variable properties of granular foods, it is challenging to manually define effective control strategies with the required level of adaptability and robustness.

A promising approach to avoid manual definition of controllers is Learning from Demonstration (LfD) or Imitation Learning (IL), where a robot learns to perform tasks from human demonstrations \cite{gordon2019,erickson2020}. This approach enables the robot to capture how humans interact with granular materials, which can then be used to train the robot to scoop and transport food. However, collecting high-quality human demonstrations for such tasks can be labor intensive and costly in physical environments, especially when a large amount of data is needed for training \cite{erickson2020}. Although some simulation environments have been proposed \cite{isaaclab2020, erickson2019}, they are typically limited to rigid food items and lack an intuitive interface that provides 3D visual feedback and 3D control for effective task demonstration \cite{erickson2019}.

To address the limitations of current RAF systems in handling granular food items, we present VERAGMIL, a  VR-based framework supporting high-performance simulation \cite{isaaclab2020}. Our framework provides photo-realistic and accurate environment for RAF, supporting a variety of granular food items manipulation (as shown in Figure\,\ref{fig:gran-mat}), a virtual reality interface for capturing dexterous human demonstrations, and the training of adaptive LfD and IL manipulation policies.


In summary, VERAGMIL introduces an intuitive and efficient approach for training RAF systems to manipulate granular food items. Our key contributions include:
\begin{itemize}
    \item We present a novel VR-based simulation framework for granular food handling in RAF. The framework features a high-fidelity physics simulation environment and a virtual reality interface designed to capture precise human demonstrations. The complete system is publicly available at: \href{https://github.com/AmanuelErgogo/VERAGMIL.git}{\textit{\small{https://github.com/AmanuelErgogo/VERAGMIL.git}}}.
    \item We evaluate the proposed framework by training three IL models (BC, BC-RNN, BCQ) with the acquired human demonstrations on granular scooping and transport tasks, benchmarking them against a human baseline and assessing their generalization to unseen food items.
    \item We  evaluate the influence of the demonstration acquisition setup, comparing VR-based and 3D space mouse-trained policies, showing VR-based policies achieve near-human performance and better generalization.
\end{itemize}

\section{RELATED WORK}


\subsection{Robot-Assisted Feeding (RAF) Systems}
\textit{RAF systems} have been developed to assist individuals with eating tasks. Examples of such systems include commercially available solutions like Obi \cite{Obi2023}, Neater Eater \cite{NeaterEater2023}, and Bestic \cite{Bestic2023}. 
In addition, there has been ongoing research in handling food items \cite{erickson2019,sundaresan2022,parikh2023}, user preferences and autonomy, which explores how users prefer to control aspects of the feeding process, such as bite timing and speed \cite{gordon2019,bhattacharjee2020,ondras2022}, anomaly detection and safety \cite{park2017,park2019,gordon2020}, and trust, acceptance, and ethics, which explores how users perceive these systems and addresses concerns related to privacy, trust, and ethics \cite{nickelsen2019}.

Current approaches in robot-assisted feeding systems employ a variety of manipulation strategies to handle various types of food, including solid, semi-solid, and soft materials \cite{grannen2022, sundaresan2022, takahashi2021}. Fixed trajectory planning is commonly used for predictable environments and standardized food items \cite{bin_picking_benchmark}, where the robot follows pre-defined paths to acquire and transport food. While effective in structured settings, this method lacks adaptability when food shape or placement is irregular, making it unsuitable for unstructured or dynamic environments.

Adaptive trajectory planning, based on reactive control \cite{sundaresan2022} and model predictive control \cite{Wang2022ModelPC, Grandia2022PerceptiveLT}, adjusts a robot's movements in real-time by responding to the dynamic properties of food items. For example, visuo-haptic skewering \cite{sundaresan2022} leverages real-time sensory feedback to adjust force application, allowing the robot to handle deformable food with varying textures. While effective for certain food types, these approaches rely heavily on accurately modeling the interaction between the robot and the food. This becomes especially challenging when dealing with granular food items exhibiting highly non-linear behaviors and unpredictable movement patterns, making them hard to model. The difficulty of capturing these complex dynamics makes it hard for robots to generalize their manipulation strategies, limiting their ability to handle novel or highly variable food items.

Deep Reinforcement Learning (Deep RL) can potentially enable autonomous learning of manipulation strategies in RAF. However, its application is constrained by safety concerns \cite{shastha2019}, sample inefficiency \cite{Alighanbari2023DeepRL}, and the difficulty of defining effective reward functions for complex tasks like food manipulation \cite{Xie2023Text2RewardRS}.

As detailed in the subsequent section, in the RAF context, the use of LfD could lead to effective manipulation strategies capable of handling foods with complex properties, such as granular ones.

\begin{figure*}[t]
    \centering    \includegraphics[width=0.9\textwidth]{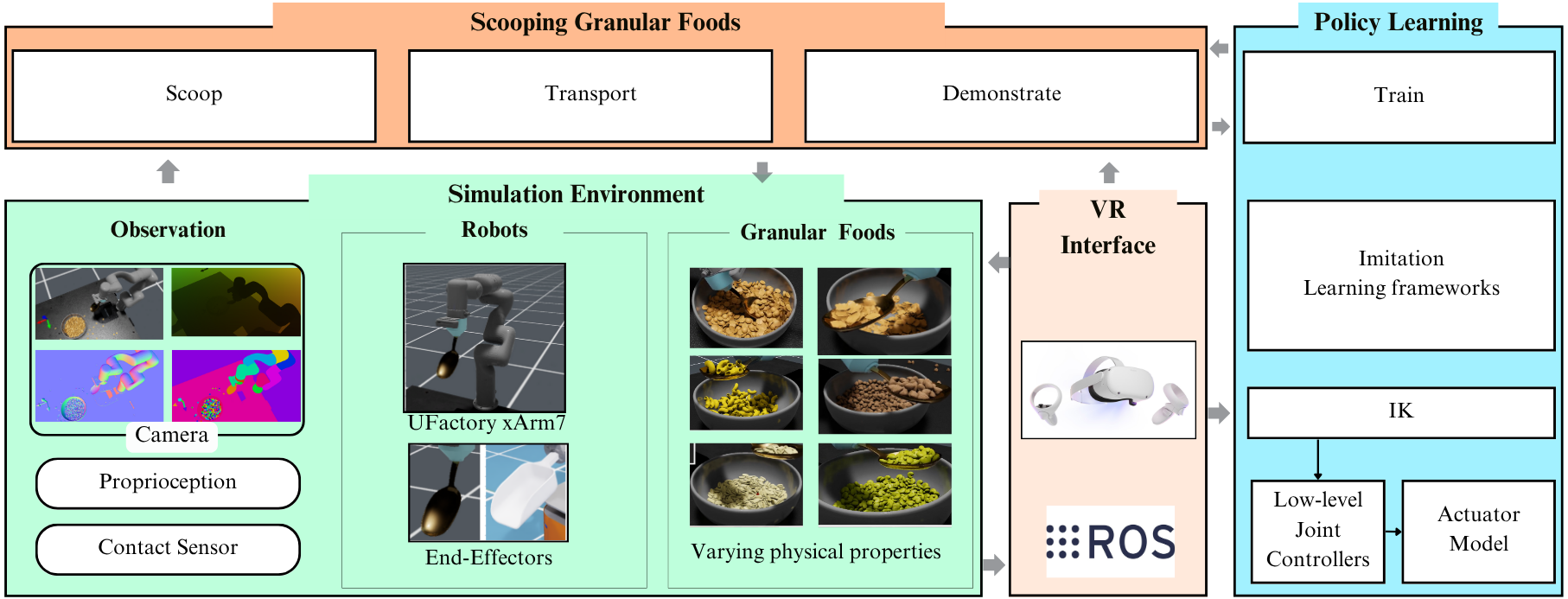}
    \caption{Overview of the VERAGMIL framework, consisting of a VR interface and a RAF simulation environment with robot platforms, granular food items, and sensors. 
    }
    \label{sim_ecosys}
\end{figure*}

\subsection{Learning from Demonstration (LfD)}
\textit{LfD} enables robots to learn complex tasks by observing human actions. In RAF, LfD helps replicate the fine motor skills required to handle diverse food textures, particularly for tasks involving granular materials where precision is crucial \cite{Sundaresan2022LearningVS, Gordon2019AdaptiveRF}. Several studies have leveraged LfD for robotic manipulation. For instance, \cite{Bhattacharjee2020IsMA} utilizes behavioral cloning to teach robots user-specific preferences, such as bite size and feeding angle, ensuring that the robot aligns with the needs of individual users. However, this approach struggles with generalizing to new, unseen situations, as it relies on directly mimicking the demonstrations without considering the underlying dynamics of the task. In addition, contextual bandit-based learning has also been used for the real-time selection of optimal feeding strategies based on the type of food and user preferences, improving adaptability \cite{Gordon2019AdaptiveRF}.

VERAGMIL supports a wide variety of LfD methods provided by the IsaacLab framework \cite{isaaclab2020}, but other frameworks can easily be supported.
This flexibility enables the comparison of different LfD methods, as detailed in this work, and facilitates the adoption of the most effective technique for the specific RAF scenario.


The effectiveness of LfD in these tasks relies heavily on the quality of the demonstrations. Since poor-quality demonstrations lead to inefficient performance of trained policies, especially in tasks that require fine motor control \cite{Tangkaratt2019VILDVI, Hao2023LeveragingDT}. Traditional methods for collecting demonstrations, such as joysticks, 3D space mice, haptic interface devices and camera-based systems, have limitations. Joysticks and haptic interfaces provide intuitive control but lack 3D visual feeback and precision, while tools like the 3D space mice offer more degrees of freedom but still require considerable operator skill \cite{Hao2023LeveragingDT}. Camera-based systems and wearable sensors offer more natural data capture but require complex acquisition setups, not always easy to integrate into the RAF scenario and which may not capture accurately fine-grained movements necessary for granular material handling\cite{Hao2023LeveragingDT}.

\textit{Virtual Reality (VR)} has emerged as an alternative approach, offering precise control in a \textit{3D virtual environment} \cite{makhataeva2020augmented, natonek1995virtual}. VR allows users to interact naturally, providing 3D visual feedback and control, enabling the capture of high-quality demonstrations that reflect the complex motions required for granular food manipulation. This approach improves the accuracy of demonstrations making it highly effective for training robots in complex tasks \cite{hilotel, badia2022virtual}. However, these existing VR solutions for human demonstrations or robot control do not provide the level of precision required for RAF tasks, are difficult to scale due to the lack of source code, the complexity of their dependencies, or the need for external workstation.

VERAGMIL overcomes these limitations, providing a ROS-based standalone VR robot interface that supports virtual and physical demonstrations, simplifies integration, and eliminates unnecessary dependencies. VERAGMIL enables realistic and natural demonstration acquisition, even in complex RAF scenarios, by fully leveraging the capabilities of VR devices. VERAGMIL also supports other demonstration acquisition interfaces, such as 3D space mice or gamepads.



\subsection{Simulation of Granular Materials}
Existing granular material simulators focus on accurate constitutive models rather than real-time robotic training, making them less suitable for the RAF context. While frameworks like \textit{Assistive Gym} and \textit{Assistive VR Gym} provide physics-based simulations for assistive tasks such as feeding, they lack \textit{GPU acceleration}, which is essential for simulating \textit{granular materials} like rice or cereals that require real-time adaptation \cite{erickson2020assistive, erickson2020vr}. Without GPU acceleration, simulations must simplify item count, interactions, and rendering realism, limiting the quality of data collected for granular interactions and reducing their effectiveness for RAF.



VERAGMIL offers a GPU-accelerated high-fidelity physics simulation that can accurately simulate the operating conditions of a RAF system. The proposed approach is based on IsaacSim which ensures the modularity and extensibility of the proposed framework \cite{isaaclab2020}. While IsaacSim supports the simulation of granular materials, we provide an integrated environment that supports a large set of granular foods and combines all the essential components, such as sensors, robots and end-effectors, tailored for the RAF application.

\section{SYSTEM DESIGN AND METHODOLOGY}

VERAGMIL addresses the complexities of RAF, particularly for granular food manipulation, by combining a high-performance simulation environment with a VR interface. This system facilitates human demonstrations and imitation learning model training for efficient granular food handling tasks. As shown in Figure \ref{sim_ecosys}, the system consists of three key components: the simulation environment (described in Section\,\ref{sec:sim-a}), the VR control interface (see Section\,\ref{sec:vrint-b}), and imitation learning frameworks (detailed in Section\,\ref{sec:lfd-c}). 

\subsection{Simulation Environment} \label{sec:sim-a}
The VERAGMIL simulation environment is developed as an Omniverse extension, so as to exploit the IsaacLab framework that provides
high-fidelity, GPU-accelerated simulations and robust frameworks for RAF \cite{isaaclab2020}. As illustrated in Figure\,\ref{sim_ecosys}, VERAGMIL framework supports various robots, end-effectors, granular food items and sensors.

\subsubsection{Granular Food Items}

As shown in Figure\,\ref{fig:gran-mat}, the dynamic behavior of different granular food items is simulated using the NVIDIA \textit{PhysX 5} physics engine, providing GPU-accelerated real-time simulation. We modeled the granular material using a set of rigid bodies implemented with the Position Based Dynamics (PBD) approach \cite{RigidBodies}. This allows granular items, such as rice and peas, to respond naturally to external forces during manipulation tasks like scooping.

As shown in Table\,\ref{table:food items}, each granular material is modeled with carefully selected physical parameters, such as shape, size distribution, density, and both static and dynamic friction coefficients, as well as restitution \cite{macklin2016xpbd}. These parameters, described with Universal Scene Description (USD) format,  are crucial for accurately replicating the behavior of these materials during RAF tasks, such as scooping and transporting \cite{dall2024ff}. For instance, foods with smaller sizes and less friction, such as rice, require more precise control and force application to minimize spillage. In contrast, larger items like beans or peas can be manipulated more easily but still demand adaptive control due to their tendency to roll or shift. To ensure realistic interactions, we use Signed Distance Field (SDF) mesh colliders tailored to each food item's geometry, enabling accurate interactions with the robot’s end effector \cite{sdf}. By capturing the distinct characteristics of each granular item, the simulation environment allows the robot to adapt its scooping and transporting strategies based on the food's behavior, improving overall task performance in RAF.

\begin{table}[tb]
\centering
\caption{Physical Properties of Food Items}
\label{table:food items}
\begin{tabular}{l|l|c|c|c|c}
\hline
\textbf{Food} & \textbf{Shape} & \textbf{Size} & \textbf{Friction} & \textbf{Restitu-} & \textbf{Density } \\ 
&  & \textbf{(cm)} & \textbf{Coef. } & \textbf{tion} & \textbf{(g/cm³)} \\ \hline
Rice                   & Elongated     & 0.5-0.7     & 0.2  & 0.1  & 1.45  \\ 
Beans                  & Oval          & 1.0-1.8     & 0.25 & 0.2  & 1.0   \\ 
Buckwheat              & Triangular    & 0.4-0.6     & 0.3  & 0.1  & 0.78  \\ 
Peas                   & Spherical     & 0.6-0.8     & 0.22 & 0.2  & 0.72  \\ 
Barley                 & Oval          & 0.8-1.2     & 0.18 & 0.15 & 1.25  \\ 
Flakes                 & Irregular     & 1.2-2.0     & 0.15 & 0.05 & 0.4   \\ 
Pasta                  & Cylindrical   & 2.0-3.5     & 0.25 & 0.1  & 1.2   \\ 
\end{tabular}
\end{table}

\subsubsection{Robotic Platforms}
The VERAGMIL framework supports different robotic systems , including the \textit{Ufactory xArm7} and \textit{Unitree G1 humanoid}, and customizable end-effectors to address various tasks in the RAF pipeline. 

In our work, we consider the setup shown in Figure\,\ref{fig:exp-setup}, where the xArm7 robotic manipulator is equipped with a standard spoon as end-effector for the manipulation of granular food. 
In this scenario, the 6-DoF control command coming from the VR interface (see Section\,\ref{sec:vrint-b})
is used to calculate desired end-effector at 60Hz and transformed into 7-DoF joint commands via task-space control using inverse kinematics \cite{isaaclab2020}. A joint-space PD controller operating at 1000Hz then processes these commands to control the torque at each joint. This high-frequency control approach  through the VR interface ensures smooth and precise manipulation of granular food items.
\begin{figure}[t]
    \centering
    \includegraphics[width=\columnwidth]{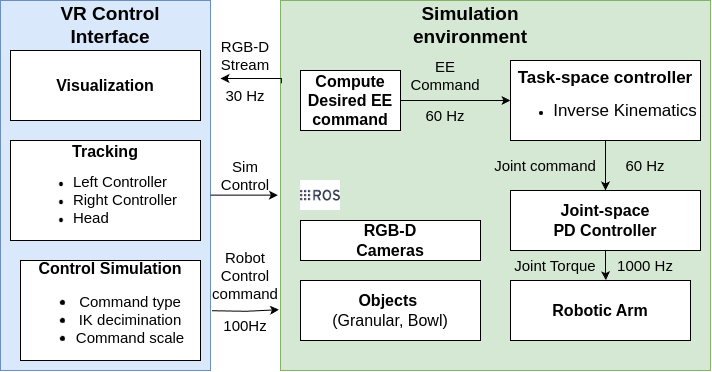}
    \caption{System architecture showing the VR interface, simulation environment and the proposed control flow.}
    \label{fig:system_diagram}
\end{figure}

\subsection{VR Interface} \label{sec:vrint-b}
The \textit{VR Interface} allows capturing high-quality human demonstrations of food handling tasks, such as scooping and transporting granular items, within a high-fidelity simulation environment. Natively supporting the Oculus Quest 2 system, the VR interface allows human operators to perform these tasks naturally and intuitively, providing an immersive experience. As shown in Figure\,\ref{fig:system_diagram}, the movements of the human demonstrator are captured and tracked using the headset’s vision system. The 6-DoF poses of both the right and left controllers, along with the headset, are transformed into the simulation’s global frame, filtered to remove high-frequency jerks, and transmitted to the simulation environment at \textit{100 Hz} to compute the desired end-effector pose.

In addition to tracking, the interface streams RGB-D data from the simulation at \textit{30Hz}, providing real-time visual feedback to the operator, essential for adjusting actions while manipulating granular materials. The interface also allows the operator to control the simulation environment, including selecting relative or absolute control commands, adjusting command frequency, and scaling to ensure synchronization with the simulation’s stepping speed, while inverse kinematics decimation ensures smooth control. 

\subsection{Data Collection and Training} \label{sec:lfd-c}
We collect demonstrations for \textit{scooping and transporting granular food items} using two interfaces: the proposed VR-based interface for natural and precise control and a 3D space mouse for comparison \cite{robosuite2020}. The system captures multi-modal data to train imitation learning models.

Each demonstration contains a recording of both proprioceptive and exteroceptive data, including the robot’s 7-DoF joint positions ($s_{joint}$), end-effector pose ($s_{ee}$), and desired post-scoop pose ($s_{desired}$). In addition, it captures RGB-D images ($camera_{rgbd}$) from a fixed overhead camera and relative movements of the end effector ($a_{ee}$) to ensure accurate task execution.

In both acquisition setups used for demonstrations, namely VR-based and 3D space mouse, we consider four different types of granular foods, i.e. rice, buckwheat, barley, and pasta. Each demonstration lasts 14-19 seconds (840-1,140 time steps at 60Hz). We collect 30 demonstrations per food item (120 per dataset) from a single human demonstrator, totaling 28-38 minutes, with overall collection spanning 42-76 minutes including resets. 

Given a collected observation-action pair dataset $\mathcal{D} = \{(s_{\text{joint}}, s_{\text{ee}}, s_{\text{desired}}, camera_{\text{rgbd}}), (a_{\text{ee}})\}$. The objective of IL is to find a policy \( \pi(a|s) \) that minimizes the loss between the demonstrated actions and the predicted actions by the model. 
For a comprehensive evaluation of imitation learning methods in RAF, we selected three methods: 
Behavioral Cloning (BC) \cite{pomerleau1988alvinn}, Recurrent Neural Networks (BC-RNN) \cite{robomimic2021}, and Batch-Constrained Q-learning (BCQ) \cite{fujimoto2019off} which are designed to handle varying levels of complexity in the dataset. In BC, the robot learns the policy directly by minimizing the loss:
\begin{equation}
\mathcal{L}(\pi) = \mathbb{E}_{(s, a) \sim \mathcal{D}} \left[ \lVert \pi(s) - a \rVert^2 \right]
\end{equation}
where \( \mathcal{D} \) is the demonstration dataset, and \( a \) represents the demonstrated actions. BC, while straightforward, suffers from compounding errors in dynamic environments like granular food manipulation. 

To handle this, BC-RNN extends BC by incorporating a recurrent structure, modeling the temporal dependencies between successive observations. The policy now accounts for a sequence of states \( s_t, s_{t-1}, \dots, s_{t-n} \), where \( n \) is the history length, allowing it to generate actions based on past observations:
\begin{equation}
\pi(a_t|s_t, h_{t-1}) = f(s_t, h_{t-1})
\end{equation}
where \( h_{t-1} \) is the hidden state from the previous timestep, allowing the model to account for time dependencies critical in continuous tasks.

BCQ introduces a reinforcement learning-based framework, where the robot learns not only to imitate but also to optimize actions based on future rewards. BCQ incorporates a Q-function \( Q(s, a) \) that estimates the future reward of an action, enabling the policy to take actions that maximize the Q-value:
\begin{equation}
\pi(a|s) = \arg\max_a Q(s, a)
\end{equation}

By integrating these three methods, our framework evaluates how effectively each method learns policies for granular food handling, comparing their adaptability and performance in dynamic environments.


\section{EXPERIMENTAL EVALUATION}
We assess the performance of the three considered imitation learning models, BC, BC-RNN, and BCQ, on granular material manipulation tasks in our RAF environment. In addition, we compare these models against a human operator baseline to further validate the learned policies. The tasks involve the scooping and transporting of granular materials such as rice, buckwheat, barley, and pasta, each presenting distinct physical properties such as size, shape, friction, and density, as reported in Table\,\ref{table:food items}. These selected granular foods provide the basis for assessing the adaptation capabilities of the considered imitation learning methods, as well as showing the performance gap between the learned policies and human dexterity.

\subsection{Experimental Setup}
\begin{figure}
    \centering
    \includegraphics[width=0.85\columnwidth]{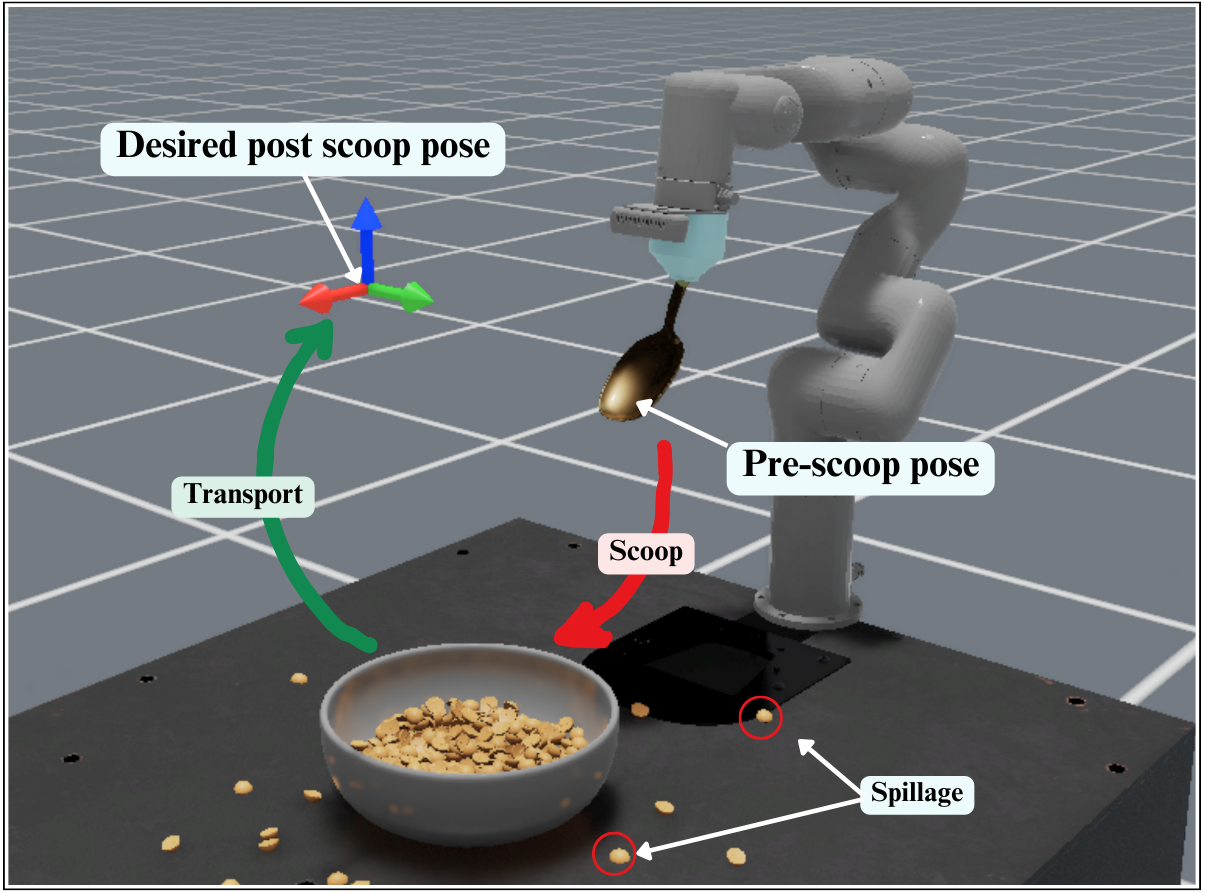}
    \caption{Experimental setup for evaluating VERAGMIL. The xArm7 manipulator robot scoops and transports granular food items, moving from the pre-scoop pose to the desired post-scoop pose, while monitoring spillage.}
    \label{fig:exp-setup}
\end{figure}

The setup involves an xArm7 robot arm with a custom spoon end effector designed to handle granular food items, as visible in Figure\,\ref{fig:exp-setup}. The curvature and size of the spoon were optimized to reflect the standard tools used in real world feeding tasks. The robot’s task is to scoop and transport food items from a container to a post-scoop pose without spilling. We selected rice, buckwheat, barley, pasta, beans, and flakes for their distinct flow, rolling, and scattering properties. This provided a comprehensive evaluation of the robot’s ability to adapt to different material behaviors. 

As described in Section\,\ref{sec:lfd-c}, each model was trained on two datasets consisting of 30 expert demonstrations for each food item (120 per dataset). 
In the first scenario, human operators controlled the robot's end effector through the VR Interface (VRI), providing natural and intuitive demonstrations of the scooping and transporting tasks. In the second scenario, demonstration data were collected using a 3D space Mouse (3DM), allowing for a comparison of the impact of different input devices on training performance. 

The food items used in the demonstrations included rice, buckwheat, barley, and pasta, representing a range of granular materials with varying physical properties. Meanwhile, peas, beans, and flakes were excluded from the training dataset and reserved for evaluating the generalization capability of the models on unseen materials.

All experiments were performed on a computer with 16GB RAM, an Intel Core i7-14700K CPU, and an RTX 4060 GPU. This hardware configuration does not impose particularly demanding specifications, demonstrating how VERAGMIL does not introduce entry barriers for the development of advanced RAF systems.


\subsection{Evaluation Metrics}
The performance of each model was evaluated based on four metrics: success rate, spillage, generalization, and task completion time. Success rate was defined as the percentage of tasks completed with less than 5\% of the total volume spilled, ensuring acceptable precision in the manipulation process. Spillage was measured by the volume of food lost, approximated as the product of the number of food pieces lost during the task and the size of each lost food item. Generalization was assessed by testing the models on unseen food items—such as peas, beans, and flakes—that were not part of the training dataset, to determine how well the models could handle new materials. Finally, task completion time was measured as the duration from the robot’s initial pre-scoop pose to the post-scoop pose (see Figure\,\ref{fig:exp-setup}), with the task considered complete when the robot successfully transferred the material while spilling less than 5\% of the total volume. These metrics provided a comprehensive evaluation of the models' performance in terms of accuracy, adaptability, and efficiency.

\begin{table}[tb]
\centering
\caption{Success rates for the different materials (\%). VRI refers to demonstrations collected in virtual reality, and 3DM refers to demonstrations using a 3D space mouse.}
\label{table:sucss-rate}
\resizebox{1\columnwidth}{!}{%
\begin{tabular}{ll|cc|cc|cc|cc}
\hline
& & \multicolumn{2}{c|}{\textbf{BC}} & \multicolumn{2}{c|}{\textbf{BC-RNN}} & \multicolumn{2}{c|}{\textbf{BCQ}} & \multicolumn{2}{c}{\textbf{Human}} \\
& \textbf{Materials} & \textbf{VRI} & \textbf{3DM} & \textbf{VRI} & \textbf{3DM} & \textbf{VRI} & \textbf{3DM} & \textbf{VRI} & \textbf{3DM} \\ 
\hline
\parbox[t]{2mm}{\multirow{4}{*}{\rotatebox[origin=c]{90}{\textbf{Training}}}}
& Rice      & 82\% & 65\% & 90\% & 75\% & 93\% & 85\% & 97\% & 91\% \\
& Buckwheat & 80\% & 60\% & 88\% & 72\% & 91\% & 80\% & 95\% & 88\% \\
& Barley    & 78\% & 58\% & 85\% & 70\% & 89\% & 78\% & 96\% & 90\% \\
& Pasta     & 65\% & 45\% & 75\% & 60\% & 80\% & 70\% & 90\% & 85\% \\
\hline\hline
\parbox[t]{2mm}{\multirow{3}{*}{\rotatebox[origin=c]{90}{\textbf{Unseen}}}}
& Peas      & 55\% & 40\% & 65\% & 50\% & 75\% & 65\% & – & – \\
& Beans     & 50\% & 38\% & 60\% & 48\% & 70\% & 60\% & – & – \\
& Flakes    & 35\% & 20\% & 50\% & 30\% & 60\% & 40\% & – & – \\
\hline
\multicolumn{2}{c|}{All Materials (avg)}
            & 64\% & 47\% & 73\% & 58\% & 80\% & 68\% & 95\% & 89\% \\
\hline
\end{tabular}
}
\end{table}

\section{RESULTS}

The results in Table\,\ref{table:sucss-rate} show that BCQ consistently outperformed both BC and BC-RNN across all scenarios. In the VRI scenario, BCQ achieved an average success rate of over 88\% on food items seen during training and more than 68\% on unseen items, approaching the human success rate of 95\% and demonstrating strong generalization capabilities. In comparison, while BC-RNN performed well on seen items with an average success rate of 84.5\%, it experienced performance drop when applied to unseen items, with an average decrease of approximately 30.95\%. BC, although performing reasonably well on seen items with an average success rate of 76.25\%, exhibited decline in performance on unseen items, achieving only a 40\% success rate in the 3DM scenario. This highlights the superior adaptability of BCQ, particularly in handling unseen food items.

Considering the results in terms of spillage, as shown in Table\,\ref{table:spillage}, BCQ outperformed BC and BC-RNN, demonstrating a 20\% reduction in spillage across all materials in the VRI scenario which is nearly three times the spillage of human demonstrations. This was particularly evident with larger items such as pasta, where BCQ’s ability to dynamically adjust actions based on material properties resulted in significantly lower spillage rates. Last row of Table\,\ref{table:spillage} further highlights BCQ’s stable task completion times across all materials in the VRI scenario, completing tasks slightly slower than BC but with considerably less spillage.

In contrast, the 3DM scenario, detailed in Table\,\ref{table:spillage}, showed an increase in spillage across all models. The challenge of precise manipulation using the 3D space mouse was evident, with BC experiencing the largest increase in spillage. Although task completion times were generally faster for all models in the 3DM scenario, this was accompanied by a noticeable decline in precision, particularly for BC, where spillage increased significantly.

Across both VRI and 3DM scenarios, as shown in Table\,\ref{table:sucss-rate}, VR-based demonstrations consistently resulted in higher success rates and lower spillage. Specifically, BCQ demonstrated a 15\% improvement in success rate and a 10\% reduction in spillage when trained on VR-based demonstrations compared to data collected using the 3D space mouse. VR-based training improved all models, with BC—previously the weakest in 3DM—gaining 16\% success, just 4\% below the best-performing model, BCQ-3DM 68\%, and 25\% below human 3DM performance 89\%. This trend, evident across all models, underscores the effectiveness of VR as a demonstration tool, as it captures more natural and intuitive human motions, leading to improved learning outcomes for the robot.

Overall, BCQ excelled across all metrics, particularly in VRI, demonstrating strong generalization. VR-based demonstrations consistently improved task success and precision, reinforcing the importance of intuitive interface. Further details on the properties of the materials and experimental setup are provided in the code repository.



\begin{table}[tb]
\centering
\caption{Spillage (Average Number of Pieces Spilled) and Task Completion Time (seconds) on Familiar Materials. VRI refers to demonstrations collected in virtual reality, and 3DM refers to demonstrations using a 3D space mouse.}
\label{table:spillage}
\resizebox{\columnwidth}{!}{%
\begin{tabular}{l|cc|cc|cc|cc}
\hline
\textbf{Metrics} 
  & \multicolumn{2}{c|}{\textbf{BC}} 
  & \multicolumn{2}{c|}{\textbf{BC-RNN}} 
  & \multicolumn{2}{c|}{\textbf{BCQ}}
  & \multicolumn{2}{c}{\textbf{Human}} \\
 & \textbf{VRI} & \textbf{3DM} 
 & \textbf{VRI} & \textbf{3DM} 
 & \textbf{VRI} & \textbf{3DM} 
 & \textbf{VRI} & \textbf{3DM} \\
\hline
Spillage (pieces)
  & 20  & 45 
  & 12  & 25 
  & 8   & 15 
  & 3   & 10 \\
Completion Time (s)
  & 20  & 30 
  & 18  & 26 
  & 15  & 22 
  & 14  & 19 \\
\hline
\end{tabular}%
}
\end{table}

\section{DISCUSSIONS}
The results demonstrate the impact of different imitation learning models and demonstration methods on the performance of granular food item handling tasks. Across both VRI and 3DM scenarios, BCQ consistently outperformed BC and BC-RNN in terms of success rate, spillage, and generalization. This is in line with previously published results that consider different robotic applications and require simpler interaction with the environment \cite{robosuite2020,robomimic2021}. We plan to extend our validation to additional imitation learning methods and also adopt demonstration-free reinforcement learning techniques to understand which solution is more effective for RAF.

The VR-based demonstrations proved to be a more effective data collection method, resulting in higher success rates and reduced spillage across all models. The VR-trained models exhibited improved adaptability to unseen granular materials, such as peas, beans, and flakes, even for simpler models like BC. The results obtained confirm those of previous works \cite{hilotel, erickson2020vr}; but we plan to conduct an extended user study, including objective usability and workload evaluation, to more clearly confirm the effectiveness of the proposed VR interface.

The use of BC-RNN provided a notable advantage in capturing sequential dependencies in granular food manipulation. However, its overall performance was inferior to that of BCQ, particularly in generalization tasks. As expected, BC performed well on familiar food items but struggled to generalize to unseen materials, highlighting the importance of dynamic learning strategies in imitation learning. Although previous studies have focused mainly on solid foods \cite{Bhattacharjee2020IsMA} or small operating spaces \cite{Sundaresan2022LearningVS}, our results extend these findings to a more complex environment.

Furthermore, task completion times were slightly faster in the 3DM scenario compared to VRI. However, this speed came at the cost of precision, with increased spillage in the 3DM scenario, especially in BC models. This trade-off between speed and precision suggests that while both input methods are viable, VR-based demonstrations offer a more balanced approach to ensure accuracy in challenging tasks related to granular food handling. 

To make our validation even more thorough and accurate, we are actively working to extend the experimental validation to consider the real environment, adopting techniques to reduce the gap between simulation and reality \cite{scheikl2022sim,kaleta2024minimal}. We also plan to improve the simulation realism, including more granular foods and their interaction with liquids of different densities and viscosities, in order to simulate soups or foods with mixed compositions.

\section{CONCLUSION}
This paper presents VERAGMIL, a VR-based imitation learning framework for granular food manipulation in RAF, which can be adapted to other granular manipulation skills with small task-specific configurations, featuring a high-fidelity simulation environment and an intuitive VR demonstration interface. We evaluated VERAGMIL by training imitation learning models on scooping and transporting tasks, using VR-based and 3D space mouse demonstrations, and comparing them to a human-expert baseline. VERAGMIL improves success rates and reduces spillage, with models approaching human performance. Although spillage is significantly reduced, fine-grained control over small portions of granular materials requires improvement. Future work will focus on sim-to-real transfer, real-world testing, and user studies.

\balance
\bibliographystyle{IEEEtran}
\bibliography{references}
\end{document}